\documentclass[runningheads]{llncs}
\usepackage[T1]{fontenc}
\usepackage{graphicx}
\usepackage[hidelinks]{hyperref}
\usepackage[small]{caption}
\usepackage{amssymb} 
\usepackage{adjustbox}
\usepackage{xcolor,colortbl}
\usepackage[flushleft]{threeparttable}
\usepackage{booktabs}
\usepackage{algorithm2e}
\RestyleAlgo{ruled}
\usepackage{subcaption}
\usepackage[misc]{ifsym}
\begin{document}
\title{Interpretable Graph Neural Networks for Heterogeneous Tabular Data}
%
%
\author{Amr Alkhatib\orcidID{0000-0003-2745-6414}(\Letter) \and
Henrik Boström\orcidID{0000-0001-8382-0300}}
\authorrunning{A. Alkhatib and H. Boström}
%
\institute{KTH Royal Institute of Technology\\
Electrum 229, 164 40 Kista, Stockholm, Sweden\\
\email{\{alkhat,bostromh\}@kth.se}}
\maketitle              
\begin{abstract}
Many machine learning algorithms for tabular data produce black-box models, which prevent users from understanding the rationale behind the model predictions. In their unconstrained form, graph neural networks fall into this category, and they have further limited abilities to handle heterogeneous data. To overcome these limitations, an approach is proposed, called IGNH (Interpretable Graph Neural Network for Heterogeneous tabular data), which handles both categorical and numerical features, while constraining the learning process to generate exact feature attributions together with the predictions. A large-scale empirical investigation is presented, showing that the feature attributions provided by IGNH align with Shapley values that are computed post hoc. Furthermore, the results show that IGNH outperforms two powerful machine learning algorithms for tabular data, Random Forests and TabNet, while reaching a similar level of performance as XGBoost.

\keywords{Machine Learning \and Graph Neural Networks \and Explainable Machine Learning.}
\end{abstract}
\section{Introduction}\label{intro}

Tabular data are predominant in certain domains like medicine, finance, and law, where trustworthiness is a central concern. The application of machine learning models in such sensitive domains often requires communicating the reasons for a prediction in order to build trust in the deployed predictive model \cite{LakkarajuKCL17} and for justifying the decisions \cite{Goodman2017}. However, many state-of-the-art machine learning algorithms produce what essentially can be considered to be black-box models for which the reasoning behind the predictions cannot be easily communicated. Post-hoc explanation methods, e.g., SHAP \cite{SHAP} and LIME \cite{LIME}, have been proposed to explain predictions of black-box models. Nevertheless, post-hoc explanation methods come with a high computational cost and limitations in the explanations they generate \cite{alkhatib23a,jethani2022fastshap}. In many cases, the generated explanations are provided with no guarantees regarding their fidelity, i.e., there is no guarantee that the explanation accurately reflects the logic of the underlying model \cite{NEURIPS2019_a7471fdc,delaunay:hal-03133223}. For these reasons, learning inherently interpretable (white-box) models may be considered when trustworthiness is a core concern \cite{rudin2019stop}.

Algorithms that generate interpretable models, e.g., logistic regression, decision trees, and generalized linear models, can be employed to give insights into how the predictions are composed. However, in most cases, opting for white-box models may lead to a substantial degradation in predictive performance for specific problems compared to state-of-the-art black-box models \cite{Loyola}. Therefore, more attention has been directed towards providing interpretable machine learning models that retain state-of-the-art performance on tabular data, e.g., TabNet \cite{Arik_Pfister_2021} and LocalGLMnet \cite{LocalGLMnet}. 
Recently, the application of Graph Neural Networks (GNNs) as a powerful means for representation learning has been extended to tabular data. Such extension involves utilizing GNNs to capture hierarchical structures within tabular data and model interactions between features, hence enabling the acquisition of enhanced data representations. TabGNN \cite{TabGNN}, TabularNet \cite{du2021tabularnet}, and Table2Graph \cite{ijcai2022p336} are examples of approaches for applying GNNs to tabular data, where TabGNN represents tabular data instances as nodes in a graph, while TabularNet and Table2Graph learn to model the interactions between features. However, the mentioned approaches produce black-box models; therefore, their applications in real-world scenarios can be limited when trustworthiness and explainability are fundamental demands. An approach to enable interpretability while leveraging the enhanced tabular data representation of GNNs, called IGNNet, was recently proposed in \cite{alkhatib2023interpretable}, by which interactions between features were modeled by representing the data points as graphs. 
However, similar to conventional deep learning algorithms for tabular data, heterogeneous data, i.e., consisting of numerical, categorical, and missing values, are problematic \cite{Borisov2023,dl_Shwartz}, and IGNNet falls short of adequately handling such data. Therefore, we present a GNN approach tailored for heterogeneous tabular data, capturing interactions within and across numerical and categorical features seamlessly, while yielding explanations together with the predictions.

The main contributions of this study are:

\begin{itemize}
    \item an approach, called IGNH (\emph{I}nterpretable \emph{G}raph Neural \emph{N}etwork for \\ \emph{H}eterogeneous tabular data), which handles both categorical and numerical features, and jointly with each prediction provides an exact feature attribution, i.e., the sum of the output feature weights gives the prediction

    \item a large-scale empirical investigation, demonstrating that the feature attributions of IGNH are the same as the Shapley attributions computed using a post-hoc explanation method
    
    \item the empirical investigation also shows that the predictive performance of IGNH is comparable to state-of-the-art approaches for tabular data; it is shown to significantly outperform Random Forests and TabNet, while reaching the performance level of XGBoost.

\end{itemize}

The next section provides an overview of related work. Section \ref{method} describes the proposed approach for categorical tabular data using graph neural networks. Moving on to Section \ref{Evaluation}, we will present and discuss the outcome of a large-scale empirical investigation. 
Finally, Section \ref{CR} summarizes the key findings and points out directions for future work.

\section{Related Work}\label{related_work}

The purpose of the proposed approach is not to provide a method for explaining black-box GNNs but to provide transparent models with high predictive performance for tabular data that can process categorical or numerical features seamlessly using GNNs. Therefore, we focus on the approaches that provide inherently interpretable GNNs. We start with a brief review of approaches to provide interpretable machine learning models based on GNNs. Afterwards, we summarize IGNNet.

The Self-Explaining Graph Neural Network (SE-GNN) \cite{SEGNN.3482306} is an explainable GNN for node classification that employs node similarities for predicting node labels, and the explanations are obtained through the K most similar nodes with labels. Kernel Graph Neural Network (KerGNN) \cite{KerGNNs} enhances the model interpretability by showing a common graph structure (graph filter) within a dataset that contains helpful information showing the structural characteristics of the dataset. ProtGNN \cite{Zhang2022ProtGNNTS} is an interpretable graph neural network that measures similarities between the input graph and prototypical patterns learned for each class, and the explanations are generated through case-based reasoning. Xuanyuan et al. \cite{Xuanyuan_} analyzed individual GNN neurons' behavior and proposed producing global explanations for GNNs using neuron-level concepts to give users a high-level view of the trained GNN model. Cui et al. \cite{Interpretable_GNN_Cui} introduced a framework that constructs interpretable GNNs for analyzing brain disorders based on connectome data. The framework models structural and functional mechanisms inside the human brain and resembles the signal correlation observed between distinct brain regions. Such approaches that promote the explainability of GNNs do not allow the users to follow the exact computations and provide abstract views of how the predictions are formulated.

IGNNet \cite{alkhatib2023interpretable} provides interpretable GNN-based models for tabular data. The tabular data instances are represented as graphs, where the nodes are the features of the instance, and the edges are the correlation between features. IGNNet maintains the interpretability of the GNN model by implementing four conditions: i) using a distinct node per feature across layers, ii) binding each node to interact with a particular neighborhood, where it sustains correlations with the nodes of the neighborhood, and the information gathering is based on linear relationships, iii) self-loop with high weight to keep a connection between the input node features and the updated node representation, and iv) an injective mapping style readout function that allows the contribution of each node to be directly linked to the predicted outcome.
    
Neverthless, IGNNet is designed mainly for predominantly numerical data and categorical features are problematic for the following reasons: first, IGNNet relies on Pearson correlation to compute the correlation between features, which proves inadequate when handling correlation involving both categorical and numerical features or between two categorical features. Second, IGNNet employs one-hot-encoding to represent categorical features, resulting in a significant rise in dimensionality when dealing with categorical features encompassing numerous categories. There is also the problem of the null graph (a completely disconnected graph) when a dataset has very weak or no correlations between features. The null graph can also occur because IGNNet establishes a threshold for a correlation value to be considered an edge in the graph.

\section{The Proposed Method}
\label{method}

The proposed method, IGNH, addresses both categorical and numerical features properly and handles categorical features differently from numerical ones. At the same time, IGNH maintains the conditions of the approach in \cite{alkhatib2023interpretable}, i.e., the message-passing layers represent each feature in a distinct node, the correlation between features bounds the interactions between nodes, weighted self-loops are added, and an injective readout function that allows tracking the contribution of each node. 

In the following subsections, we start by describing the data preprocessing necessary to convert tabular data instances into graphical representations. Subsequently, we outline the training process of the proposed method as a straightforward GNN for graph classification. Finally, we describe how IGNH works at inference time and the derivation of local (instance-based) explanations.

\subsection{Data Preprocessing}\label{preprocessing}

The preprocessing step is done once, before the training phase, using the training data examples, and the resulting graph structure is shared between the training data instances and maintained for the data examples at inference time. 

In order to present a tabular data instance as a graph, each feature is initially represented by a node of one-dimensional feature vector with a value equal to the feature value. To determine the presence of an edge between two nodes, the correlation between the two features they represent is measured and incorporated into the graph as an edge weight. Consequently, the resulting graph is weighted, with edge weights ranging from -1 to 1. Measuring the linear relationships between features, such as through the Pearson coefficient \cite{Pearson_corr}, can be beneficial in modeling the interactions between features while contributing to the overall interpretability of the model. Pearson correlation measures the linear relationship between two variables as depicted in Formula \ref{pearson}. However, categorical variables are qualitative, i.e., such a linear relationship cannot be computed using Formula \ref{pearson}. Therefore, the Pearson coefficient falls short of adequately measuring the correlation between numerical and categorical features and between two categorical features. 

\begin{equation}\label{pearson}
r = \frac{\sum_{i=1}^{n} (x_i - \mu_{x}) (y_i - \mu_{y})}{\sqrt{\sum_{i=1}^{n}(x_i - \mu_{x})^2 \sum_{i=1}^{n}(y_i - \mu_{y})^2}}
\end{equation}

\noindent Where $\mu_{x}$ and $\mu_{y}$ are the means of the variable x and variable y respectively, and $n$ is the sample size.

There are two cases of correlation between features that cannot be handeled adequately by Pearson coefficient: correlation between categorical and numerical features, and correlation between two categorical features. We propose the use of a different correlation function for each case. Subsequently, in order to measure the association between categorical and numerical variables, we propose using the Point-biserial correlation coefficient \cite{pointbiserialr}, a special case of the Pearson correlation adapted to handle one dichotomous (binary) variable. Therefore, if $X$ is a continuous variable and $Y$ is a binary variable, the Point-biserial correlation is computed as follows:

\begin{equation}\label{biserial}
r = \frac{\mu_{1} - \mu_{0}}{\sigma_n} \sqrt{\frac{n_1 n_0}{n^2}}
\end{equation}

\noindent Where $\mu_{1}$ is the mean of $X$ when $Y$ is observed 1, $\mu_{0}$ is the mean when $Y$ is zero, and $\sigma_n$ is the standard deviation of $X$. $n_1$ is the sample size when $Y$ is 1, $n_0$ is the sample size when $Y$ is zero, and $n$ is the total sample size.

Consequently, for the sole objective of computing the correlation between categorical and numerical features, we binarize categorical features using the one-hot-encoding and compute the Point-biserial correlation between the numerical feature and each binarized category. In order to derive a single value representing the relationship between a numerical feature and a categorical feature, we apply the Fisher's z correction of Corey et al. \cite{averaging_correlations}. First, we apply Fisher's transformation \cite{FisherTransformation} to the correlation ($r$) values computed between the numerical feature and each category. Afterwards, the transformed values are averaged, and the transformation is reversed, converting the averaged value back to a correlation ($r$) using Fisher's inverse transformation.

Finally, ordinal associations can be more meaningful in modeling interactions between categorical variables than linear relationships. Therefore, the Kendall rank correlation coefficient (Kendall's $\tau$ coefficient) \cite{kendalltau} is employed for the correlation between categorical features, which measures the ordinal association between two quantified variables, as shown in Formula \ref{Kendall}. 

\begin{equation}\label{Kendall}
\tau = 1 - \frac{2 \Delta_d(\mathbb{X}, \mathbb{Y})}{n(n - 1)}
\end{equation}

\noindent Where $\Delta_d$ is the difference between the number of concordant and discordant pairs of sets $\mathbb{X}$ and $\mathbb{Y}$.

In order for a correlation value between two variables to qualify for a weighted edge in the graph, it has to satisfy a statistical significance test, which eliminates irrelevant correlation values and restrains the interactions of a node to a neighborhood with relevant connections. Moreover, the significance test evades building a fully connected graph with numerous noisy correlations (edge weights). Therefore, we test the null hypothesis for each correlation value that the association between features is absent, i.e., correlation = 0. Accordingly, if the p-value is below a certain statistical significance level (e.g., 0.05 or 0.01), the correlation value is included as an edge in the graph and will be dismissed otherwise. For the Point-biserial correlation, only the statistically significant values are passed to the Fisher's z correction for averaging. Hence, any computed correlation value is incorporated into the graph as long as the correlation is statistically significant, which is anticipated to address the null graph issue that arises when weak correlations are encountered.

\subsection{IGNH Training}
Following the transformation of tabular data points into graphs, where features serve as nodes and computed correlation values as edges. Each feature is represented by an initial feature vector of one dimension, and the GNN will embed this initial value into higher dimensionality before the message passing layers. IGNH at the input layer addresses the categorical features differently from the numerical features, where the categorical features are passed to a learnable embedding layer that learns and stores numerical presentation for each category. Numerical features are passed to a linear transformation layer that projects the numerical feature values into the same dimensionality as the embeddings of the categorical features. Afterwards, the learned vectors of both categorical and numerical features form the node representations that are fed to the message-passing layers of the GNN, which in turn update the node representations as follows: 

\begin{equation}\label{eq_mp}
\mathbf{h}_i^{(l+1)} = \varphi\left(\mathbf{w}^{(l+1)}\left(\omega_{i,i} \mathbf{h}_i^{(l)} + \sum_{u \in \mathcal{N}(i)} \omega_{i,u} \mathbf{h}_u^{(l)}\right)\right)
\end{equation}

\noindent Where  $\textbf{h}_i^{(l)}$ is the hidden representation of the node $v_i$ in the $l^{th}$ layer, $\textbf{w}^{(l+1)}$ is the learnable parameters for each node, $\omega_{i,u}$ is the weight of the edge between node $v_i$ and node $v_u$, $\varphi$ is a non-linearity function, and $\mathcal{N}(i)$ is the neighborhood of node $v_i$.


The weight of the self-loop $\omega_{i,i}$ is essential for the model's interpretability, as it accounts for the main message carried by the node across the layers of the GNN. Therefore, we assign a high weight to the self-loop to maintain the interpretability of the model.

After the message-passing layers, the final node representations are projected into scalar values using an injective readout function, e.g., linear transformation without activation functions, that maps each node to a single dimension in the final graph representation ($\mathcal{R}(\textbf{h}_i^{(l+1)} \in \mathbb{R}^n) = h_i \in \mathbb{R}^1$), which allows tracking each (node) feature's influence on the predicted outcome. Each scalar value represents one node and, subsequently, represents one input feature. 

The learned graph representation is employed to formulate a prediction, as shown in Equation \ref{inj_readout}.

\begin{equation}\label{inj_readout}
\hat{y} = \rho \left(\sum_{i=1}^n{w_{i} \mathcal{R}\left(\textbf{h}_i^{(l+1)}\right)}\right)
\end{equation}

\noindent Where $\rho$ is a link function that accommodates a valid range of outputs, e.g., the sigmoid function for binary classification tasks, $w_i$ is the weight of node $v_i$, $n$ is the total number of the nodes, and $\mathcal{R}$ is an injective readout function.

\begin{figure*}[h]
    \centering
    \includegraphics[width=1.\textwidth]{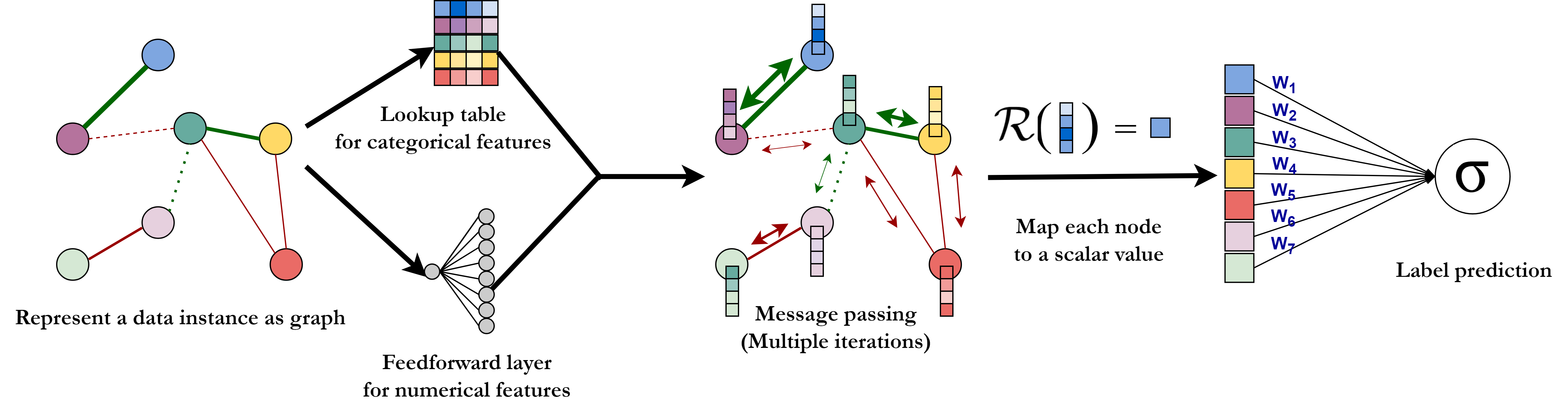}
    \caption{\textbf{An overview of the proposed approach.} Each data example is represented as a graph. The features of the data instance are the nodes and the edges between nodes are the correlation between features. Multiple iterations of message passing are applied. Finally, the obtained node representations are projected using an injective mapping function into scalar values, and the graph representation is obtained by concatenating the projected values and used for prediction.}
    \label{fig:illust1}
\end{figure*}

Finally, all the parameters $\theta$ of the GNN can be optimized using a suitable loss function, e.g., cross-entropy for classification or mean squared error of regression.
The method is illustrated in Figure \ref{fig:illust1} and summarized in Algorithm \ref{alg:1}.

\begin{algorithm}[H]
\SetAlgoLined
\small
\caption{IGNH}
\label{alg:1}
\KwData{a set of graphs $\mathbb{G}$ and labels $\mathbb{Y}$}
\KwResult{model parameters $\theta$}
Initialize $\theta$\\
\For{number of training iterations}{
$\mathcal{L} \gets 0$\\
    \For{each $\mathcal{G}_j \in \mathbb{G}$}{
    $\textit{\textbf{H}}^{(cat)}_j \gets $LookupTable$(categorical \in \mathcal{G}_j)$\\
    $\textit{\textbf{H}}^{(num)}_j \gets $Feedforward$(numerical \in \mathcal{G}_j)$\\
    
    $\textit{\textbf{H}}_j \gets \textit{\textbf{H}}^{(cat)}_j \mathbin\Vert \textit{\textbf{H}}^{(num)}_j$\\
    
        \For{each layer l $\in$ messagePassing layers}{
        $\textit{\textbf{H}}^{(l+1)}_j \gets $messagePassing$(\textit{\textbf{H}}^{(l)}_j)$\\
        }
    $\textbf{g}_j \gets $readout$(\textit{\textbf{H}}^{(l+1)}_j)$\\ 
    $\hat{y}_j \gets $predict$(\textbf{g}_j)$\\
    $\mathcal{L} \gets \mathcal{L} + $loss$(\hat{y}_j,y_j \in \mathbb{Y})$
    }
Compute gradients $\nabla_{\theta} \mathcal{L}$\\
Update $\theta \gets \theta - \nabla_{\theta} \mathcal{L}$
}
\end{algorithm}

\subsection{At Inference Time}
In order to make a prediction for a new data instance, the data example has to be transformed into a graph using the same procedure mentioned in Subsection \ref{preprocessing}, for which the graph structure obtained using the training data prior to the model training is employed, eliminating the need to recompute correlation values between features at inference time. 

Explanations can be derived from the final individual scores obtained after the readout function $w_{i}\mathcal{R}(\textbf{h}_i^{(l+1)})$, which encapsulate all computations performed by the GNN for the $i^{th}$ feature and can be interpreted similarly to the feature importance scores obtained via SHAP or LIME, and their summation provides the predicted outcome by IGNH.

\section{Empirical Investigation}
\label{Evaluation}

This section begins by describing the experimental setup, and then the explanations output by IGNH are evaluated through a comparison to the Shapley attributions. Afterwards, we benchmark the predictive performance in comparison to state-of-the-art methods for tabular data.

\subsection{Experimental Setup} \label{experimental_setup}

The investigation involves 30 publicly accessible datasets, which are partitioned into training, validation, and test sets.\footnote{All the datasets are obtained from \url{https://www.openml.org}} \footnote{Detailed information about the datasets is provided in the technical appendix.} The training set is used to train the model, the validation set is employed to detect overfitting and enforce early stopping during the training phase, and the test set is used to evaluate the model's performance. Out of the 30 datasets used in the following experiments, 13 are composed of a mixture of numerical and categorical features, 12 are numerical, and 5 are categorical. However, half of the datasets are predominantly categorical. In order to ensure a fair comparison, all the learning algorithms in the following experiments are trained without hyperparameter tuning, using default settings on each dataset. If the algorithm does not use the development set to monitor performance progress, e.g., XGBoost and Random Forests, the validation and training sets are combined in an augmented training set.

IGNH employs the same architecture and set of hyperparameters described in \cite{alkhatib2023interpretable} with 6 message-passing layers and the self-loop accounts for around 90\% of the aggregated messages.\footnote{The source code is available here: \url{https://github.com/amrmalkhatib/IGNH}} Since we rely on the statistical significance to decide which correlation value is considered an edge weight in the graph, we test the statistical significance at the 0.05 level. TabNet is trained with early stopping, triggered after 20 consecutive epochs if no improvement is observed on the validation set, and the best-performing model is selected for evaluation. Regarding the imbalanced binary classification datasets, we conduct random oversampling of the minority class in the training set, aligning its size with the majority class. All the experimented algorithms are then trained using the oversampled training data. On the other hand, multi-class datasets are not oversampled. The area under the ROC curve (AUC) is the main metric for measuring predictive performance, as it is a classification-threshold-invariant metric. A weighted AUC is computed for multi-class datasets, which involves computing the AUC for each class against the rest and weighting it by the corresponding class support.

In the initial data preprocessing stage, the categories of each categorical feature are tokenized using numbers starting from one upwards, while zero is reserved for missing values. One-hot encoding is not employed to represent categorical features since many of the used datasets comprise numerous categories, leading to high dimensionality that can exceed memory constraints. 
The numerical features are normalized using standard normalization. Normalization ensures that the feature values are constrained within the same range to maintain a consistent scale across all numerical nodes.

\subsection{Evaluation of Explanations}

In order to evaluate the explainability of IGNH predictions, we follow the approach proposed in \cite{alkhatib2023interpretable}, where the feature scores generated by IGNH are expected to accurately reflect the contribution of each feature to the predicted outcome and consequently aligned with the true Shapley values. With the aim of showing if such alignment exists, we employ KernelSHAP as it has been shown to converge to the true Shapley values given an infinite number of samples \cite{covert21a,jethani2022fastshap}. Consequently, the explanations provided by KernelSHAP are expected to gradually converge towards values more similar to the scores produced by IGNH, if these scores align with the true Shapley values.

Therefore, KernelSHAP is used to explain IGNH, and the generated explanations are compared to IGNH's scores after each iteration of KernelSHAP's data sampling and evaluation. The experiment is conducted using the 30 datasets. However, in order to conduct the experiment on 30 datasets within a reasonable timeframe, 500 examples are randomly selected from the test set of each dataset for explanation purposes. The similarity metrics between explanations are the cosine similarity and Spearman rank-order correlation. The cosine similarity measures the alignment in orientation, and the Spearman rank order measures the alignment in ranking the importance scores.

\begin{figure*}[h]

\centering
\includegraphics[width=.34\textwidth]{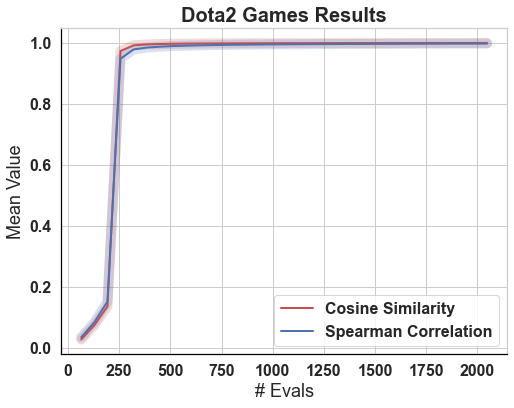}\hfill
\includegraphics[width=.33\textwidth]{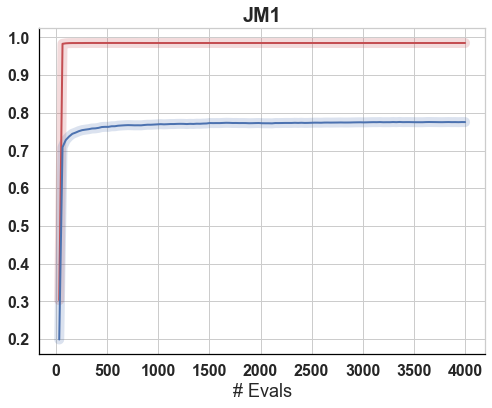}\hfill
\includegraphics[width=.33\textwidth]{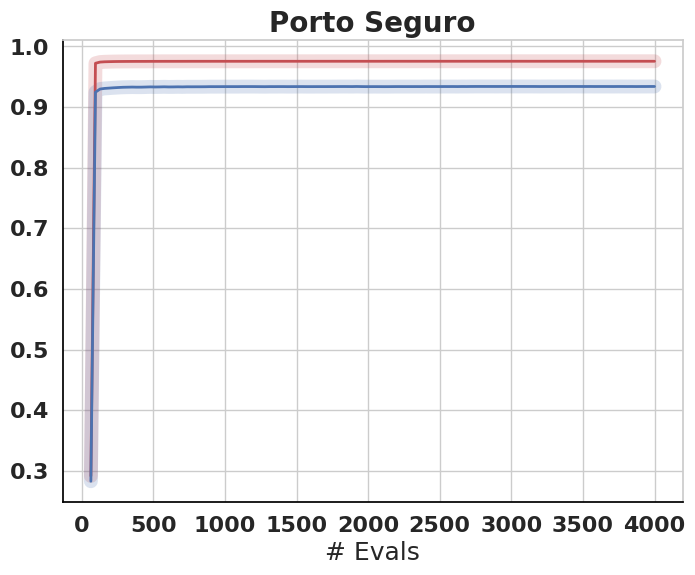}

\includegraphics[width=.34\textwidth]{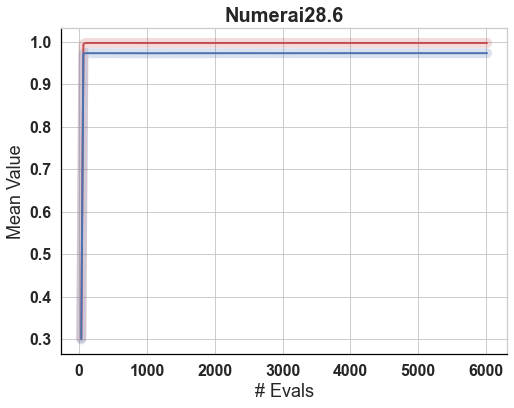}\hfill
\includegraphics[width=.33\textwidth]{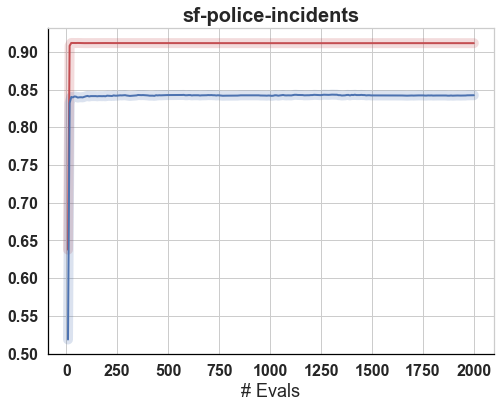}\hfill
\includegraphics[width=.33\textwidth]{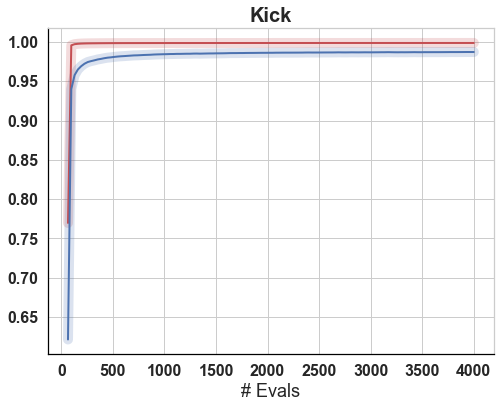}

\caption{\textbf{Comparison between the approximations generated by KernelSHAP and the importance scores obtained from IGNH.} We assess the similarity of KernelSHAP's approximations to the scores produced by IGNH during the iterations of data sampling and evaluation by KernelSHAP. It becomes evident that KernelSHAP shows improved accuracy in approximating the scores derived from IGNH with further data sampling.}
\label{fig:transparency}

\end{figure*}

The results show a consistent pattern where KernelSHAP's explanations always converge towards values more similar to IGNH's scores on different data examples and different datasets, as depicted in Figure \ref{fig:transparency}.\footnote{The complete set of results on the 30 datasets is available in the technical appendix.} This consistent convergence indicates that IGNH is providing transparent models with feature scores aligned with the true Shapley attributions. Additionally, the employment of the Kendall rank correlation and the Point-biserial correlation, besides the Pearson correlation, maintain the transparency of the learned models while properly handling the correlations between different variable types.

\subsubsection{Illustration of Explanation.}
\begin{figure}[h]
  \centering
  \subfloat[The original data point.]{\includegraphics[width=0.49\textwidth]{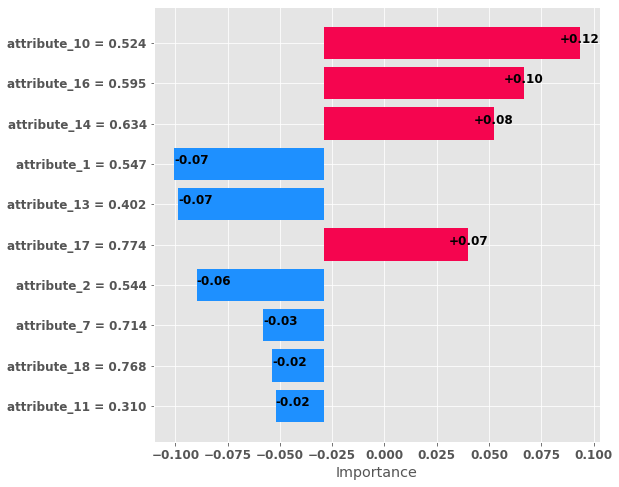}\label{fig:numerai_org}}
  \hfill
  \subfloat[The data point with a modified attribute\_10. ]{\includegraphics[width=0.49\textwidth]{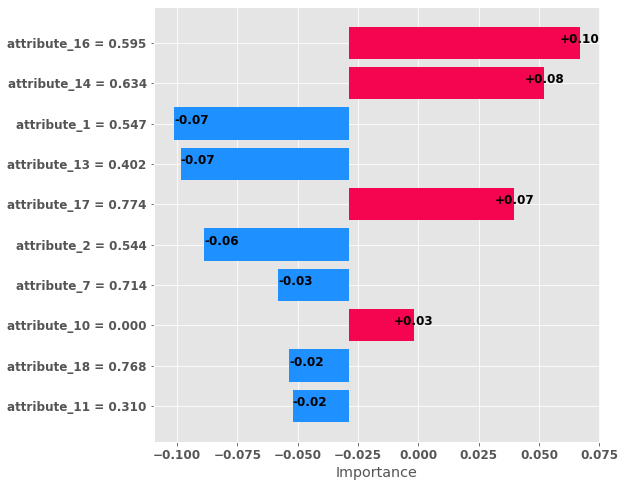}\label{fig:numerai_mod}}
  \caption{Explanation to a single prediction on the Numerai 28.6 dataset.}
  
  \label{fig:post_hoc}
\end{figure}

Using a toy example from the Numerai 28.6 dataset, we demonstrate how the feature scores computed by IGNH offer insights into the features influencing the predicted outcome. The explanation can be interpreted similarly to the explanations generated by a SHAP or LIME explainer as shown in Figure \ref{fig:post_hoc}. In our illustration, the feature scores are centered around the bias value. The displayed scores produce the prediction of IGNH when the scores are summed with the bias and the link function is applied. The scores are sorted in descending order according to their absolute values and centered around the bias value. We plot only the top 10 feature scores for clarity.
In the specific instance provided (Figure \ref{fig:numerai_org}), IGNH produced a positive prediction with a narrow margin of 0.52. To validate whether the explanation aligns with IGNH's decision-making process, we reduce the value of the top feature (attribute\_10) to zero while keeping the remaining feature values unchanged. This adjustment shifts the prediction to a negative outcome with a value of 0.497. We can also observe that the positive influence attributed to the attribute\_10 decreases from 0.12 in the original data instance to 0.03 in the modified one, as depicted in Figure \ref{fig:numerai_mod}. Consequently, the user can manipulate the predictions of IGNH based on the provided explanations.

\subsection{Evaluation of Predictive Performance}\label{performance}

In the following experiment, we use 30 datasets for evaluation. 
We evaluate the predictive performance of IGNH relative to that of XGBoost, Random Forests, and TabNet.

First, we assessed the performance of IGNH in comparison to IGNNet across 15 datasets characterized by numerical features. IGNNet is designed to handle datasets with predominantly numerical features, as categorical features with numerous categories lead to an exponential increase in dimensionality and the consequent number of nodes. The detailed results of this experiment are available in the appendix, which show that IGNH and IGNNet perform on the same level on datasets with mainly numerical features.

\begin{figure*}[h]
    \centering
    \includegraphics[width=1.\textwidth]{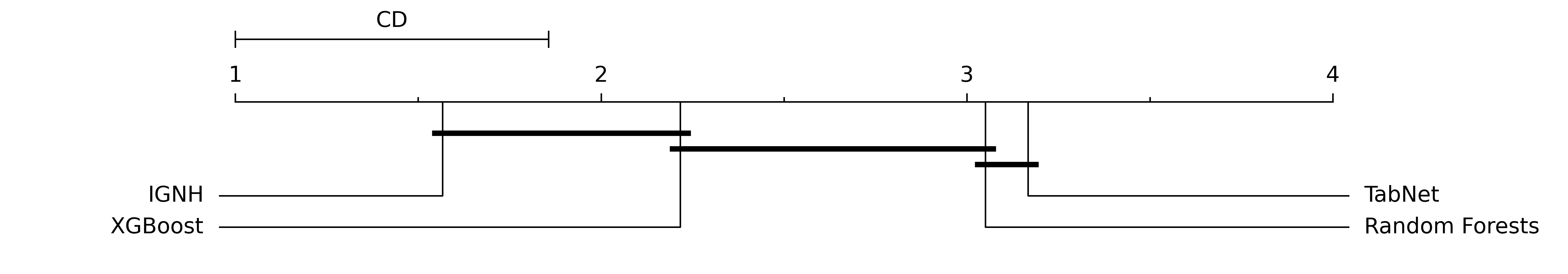}
    \caption{\textbf{The average rank of the compared algorithms over the 30 datasets with respect to the AUC}, where a lower rank is better. The critical difference (CD) shows the biggest difference that is not statistically significant.}
    \label{fig:ranks}
\end{figure*}

\subsubsection{Comparison to Powerful Algorithms.}

The results for IGNH, XGBoost, Random Forests, and TabNet across the 30 datasets are shown in Table 2. The four competing algorithms are ranked across the 30 datasets based on their predictive performance measured by the AUC metric, and the outcome demonstrates that IGNH comes in first place, followed by XGBoost. We employ the Friedman test \cite{Friedman_test} to assess the null hypothesis that there is no difference in the predictive performance, as measured by AUC, between IGNH, XGBoost, Random Forests, and TabNet, which allowed the rejection of the null hypothesis at the 0.05 level. Consequently, the post-hoc Nemenyi test \cite{nemenyi:distribution-free} is applied to determine which pairwise differences are significant at the 0.05 level. The results of applying the Nemenyi test are presented in Figure \ref{fig:ranks}, which show a significant difference in performance between IGNH and both Random Forests and TabNet. However, no significant difference in performance is observed between IGNH and XGBoost. The results indicate that IGNH maintains performance at the level of powerful algorithms for tabular data, e.g., XGBoost. However, datasets consisting solely of categorical features, such as Poker Hand, present significant challenges for traditional deep learning models as well as boosted decision trees, even though a hand-crafted set of rules can achieve perfect accuracy \cite{Arik_Pfister_2021}. Notably, IGNH surpasses XGBoost and Random Forests on the Poker Hand dataset, achieving a remarkable 99.4\% AUC. A similar observation can be seen on the Dota2 Games Results dataset, which is entirely categorical, where IGNH outperforms all the competitors.

\begin{table*}
\caption{The AUC of IGNH, TabNet, Random Forests, and XGBoost. The best-performing model is \colorbox[HTML]{A8B6E0}{colored in blue}.}
\centering
\begin{adjustbox}{width=.9\textwidth}
\small
\begin{threeparttable}
\begin{tabular}{l c c c c}
    \toprule
    \rowcolor[HTML]{EFEFEF} 
    \multicolumn{1}{c}{{\cellcolor[HTML]{EFEFEF}Dataset}} & \multicolumn{1}{l}{\cellcolor[HTML]{EFEFEF}IGNH} & TabNet & Random Forests & XGBoost \\
    \cmidrule(lr){1-5}
    Abalone & \colorbox[HTML]{A8B6E0}{0.882} & 0.877 & 0.877 & 0.871 \\
    Airlines & 0.714 & 0.7 & 0.659 & \colorbox[HTML]{A8B6E0}{0.725} \\
    Amazon employee access & 0.832 & 0.747 & \colorbox[HTML]{A8B6E0}{0.852} & 0.84 \\
    Autos & 0.965 & 0.944 & 0.946 & \colorbox[HTML]{A8B6E0}{0.972} \\
    Bank32nh & \colorbox[HTML]{A8B6E0}{0.887} & 0.86 & 0.876 & 0.879 \\
    Bank Marketing & \colorbox[HTML]{A8B6E0}{0.928} & 0.921 & 0.922 & 0.924 \\
    BNG solar flare & \colorbox[HTML]{A8B6E0}{0.977} & 0.973 & 0.931 & \colorbox[HTML]{A8B6E0}{0.977} \\
    BNG cmc nominal & \colorbox[HTML]{A8B6E0}{0.71} & 0.698 & 0.689 & 0.705 \\
    Click prediction small & 0.678 & 0.656 & 0.668 & \colorbox[HTML]{A8B6E0}{0.695} \\
    Covertype & 0.985 & 0.967 & \colorbox[HTML]{A8B6E0}{0.996} & 0.965 \\
    Credit Card Fraud Detection & \colorbox[HTML]{A8B6E0}{0.99} & 0.93 & 0.913 & 0.951 \\
    Diabetes130US & 0.691 & 0.661 & 0.672 & \colorbox[HTML]{A8B6E0}{0.694} \\
    Dota2 Games Results & \colorbox[HTML]{A8B6E0}{0.639} & 0.574 & 0.595 & 0.625 \\
    Elevators & \colorbox[HTML]{A8B6E0}{0.944} & 0.943 & 0.908 & 0.933 \\
    Fars & 0.958 & 0.954 & 0.949 & \colorbox[HTML]{A8B6E0}{0.962} \\
    HCDR Main & 0.759 & 0.754 & 0.737 & \colorbox[HTML]{A8B6E0}{0.76} \\
    Helena & 0.874 & \colorbox[HTML]{A8B6E0}{0.885} & 0.856 & 0.874 \\
    Heloc & \colorbox[HTML]{A8B6E0}{0.79} & 0.777 & 0.781 & 0.777 \\
    JM1 & 0.736 & 0.725 & \colorbox[HTML]{A8B6E0}{0.742} & 0.701 \\
    Kick & \colorbox[HTML]{A8B6E0}{0.765} & 0.746 & 0.754 & 0.76 \\
    Madelon & \colorbox[HTML]{A8B6E0}{0.914} & 0.506 & 0.792 & 0.895 \\
    MC1 & \colorbox[HTML]{A8B6E0}{0.958} & 0.933 & 0.848 & 0.939 \\
    Microaggregation2 & \colorbox[HTML]{A8B6E0}{0.775} & 0.746 & 0.761 & 0.771 \\
    Numerai28.6 & \colorbox[HTML]{A8B6E0}{0.525} & 0.518 & 0.513 & 0.508 \\
    Pokerhand & 0.994 & \colorbox[HTML]{A8B6E0}{0.997} & 0.895 & 0.907 \\
    Porto Seguro & \colorbox[HTML]{A8B6E0}{0.633} & 0.63 & 0.599 & 0.622 \\
    PC2 & \colorbox[HTML]{A8B6E0}{0.881} & 0.684 & 0.63 & 0.683 \\
    SF Police Incidents & 0.645 & 0.586 & \colorbox[HTML]{A8B6E0}{0.653} & 0.626 \\
    Traffic Violations & 0.883 & 0.8 & 0.877 & \colorbox[HTML]{A8B6E0}{0.9} \\
    Speed Dating & 0.844 & 0.778 & 0.851 & \colorbox[HTML]{A8B6E0}{0.857} \\ \bottomrule
\end{tabular}
\end{threeparttable}
\end{adjustbox}
\label{table:benchmark}
\end{table*}

\section{Concluding Remarks}
\label{CR}

We have proposed an algorithm, called IGNH, for learning graph neural networks from heterogeneous tabular data, which are constrained to output exact feature attributions along the predictions.  
IGNH adequately models the interaction between features both of different and the same types of data, i.e., whether categorical or numerical. IGNH outputs the exact feature attribution of each prediction without the computational overhead of employing a post hoc explanation technique.

Results from a large-scale empirical investigation were presented, showing that the output explanations indeed align with the true Shapley values. The predictive performance of IGNH was compared to three state-of-the-art algorithms for tabular data; XGBoost, Random Forests, and TabNet. IGNH was observed to significantly outperform both Random Forests and TabNet, while reaching a similar level of performance as XGBoost.

One direction for future work involves studying alternative approaches to modeling feature interactions other than correlation. Another direction for future work concerns extending IGNH to accommodate also non-tabular datasets, such as images and text, or even multi-modal datasets, e.g., including both tabular and non-tabular data. Finally, one important direction concerns considering user-centric, rather than just functional, assessments, such as determining the extent to which a task could be solved more effectively when users are given access to explanations in the form of feature attributions.

\clearpage
\begin{credits}
\subsubsection{\ackname} This work was partially supported by the Wallenberg AI, Autonomous Systems and Software Program (WASP) funded by the Knut and Alice Wallenberg Foundation.

\end{credits}
%
%
%
 \bibliographystyle{splncs04}
 \bibliography{ref}

\clearpage

\appendix

\section{Evaluation of Predictive Performance}\label{performance}

The results of comparing IGNetH and IGNNet on the 15 datasets are available in Table \ref{table:baseline_experiments}. The ranking of the two algorithms using the AUC on the 15 datasets shows that IGNetH slightly outperforms IGNNet. We test the null hypothesis that there is no difference in the predictive performance between IGNetH and IGNNet, as measured by AUC, using the Wilcoxon signed-rank test \cite{wilcoxon1945individual} since only two methods are compared in this experiment. The test reveals that the null hypothesis may not be rejected at the significance level of 0.05. Therefore, we conclude that IGNetH maintains the performance level of IGNNet.

\begin{table*}
\caption{The predictive of IGNetH and IGNNet measured by the AUC. The best-performing model is \colorbox[HTML]{A8B6E0}{colored in blue}.}
\centering
\begin{adjustbox}{width=.5\textwidth}
\small
\begin{threeparttable}
\begin{tabular}{l c c}
    \toprule
    \rowcolor[HTML]{EFEFEF} 
    \multicolumn{1}{c}{{\cellcolor[HTML]{EFEFEF}Dataset}} & \multicolumn{1}{l}{\cellcolor[HTML]{EFEFEF}IGNetH} & IGNNet \\
    \cmidrule(lr){1-3}
    Abalone & \colorbox[HTML]{A8B6E0}{0.883} & 0.881 \\
    Bank32nh & 0.887 & 0.887 \\
    Covertype & \colorbox[HTML]{A8B6E0}{0.986} & 0.984 \\
    Credit Card Fraud Detection & \colorbox[HTML]{A8B6E0}{0.99} & 0.987 \\
    Elevators & 0.944 & \colorbox[HTML]{A8B6E0}{0.951} \\
    Helena & 0.874 & \colorbox[HTML]{A8B6E0}{0.875} \\
    Heloc & \colorbox[HTML]{A8B6E0}{0.79} & 0.783 \\
    JM1 & 0.736 & \colorbox[HTML]{A8B6E0}{0.739} \\
    Madelon & \colorbox[HTML]{A8B6E0}{0.914} & 0.906 \\
    MC1 & \colorbox[HTML]{A8B6E0}{0.958} & 0.957 \\
    Microaggregation2 & 0.775 & \colorbox[HTML]{A8B6E0}{0.778} \\
    Numerai28.6 & 0.525 & \colorbox[HTML]{A8B6E0}{0.526} \\
    Fars & \colorbox[HTML]{A8B6E0}{0.959} & 0.956 \\
    PC2 & 0.881 & 0.881 \\
    Speed Dating & 0.843 & \colorbox[HTML]{A8B6E0}{0.853} \\ \bottomrule
\end{tabular}
\end{threeparttable}
\end{adjustbox}
\label{table:baseline_experiments}
\end{table*}

\section{Transparency Evaluation}

In the following figures \ref{fig:transparency_full1} and \ref{fig:transparency_full2}, we present the results of the comparison between the explanations obtained from IGNetH and the explanations of KernelSHAP using 30 datasets. The results show a consistent trend across the 30 datasets where the explanations obtained using KernelSHAP always converge to more similar values to the explanations of IGNetH.

\begin{figure*}[h]

\centering
\includegraphics[width=.34\textwidth]{figures/dota2-exp.png}\hfill
\includegraphics[width=.33\textwidth]{figures/jm1-exp.png}\hfill
\includegraphics[width=.33\textwidth]{figures/porto-exp.png}

\includegraphics[width=.34\textwidth]{figures/numerai-exp.png}\hfill
\includegraphics[width=.33\textwidth]{figures/SF-Police-exp.png}\hfill
\includegraphics[width=.33\textwidth]{figures/kick-exp.png}

\includegraphics[width=.34\textwidth]{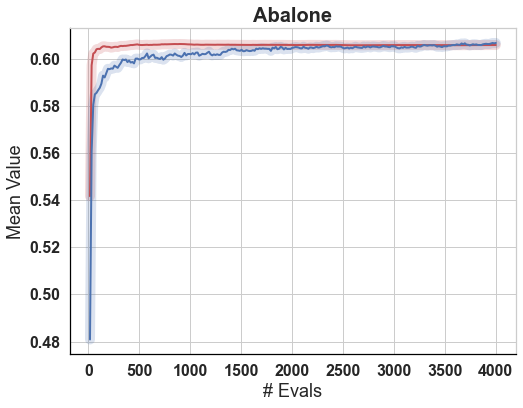}\hfill
\includegraphics[width=.33\textwidth]{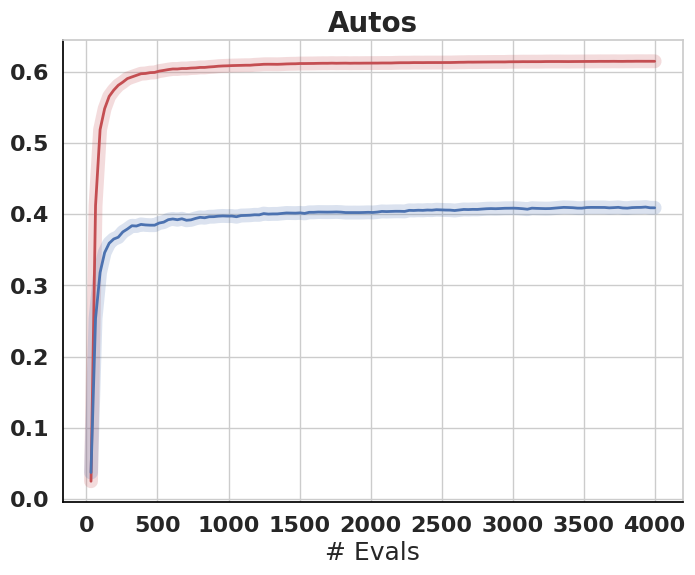}\hfill
\includegraphics[width=.33\textwidth]{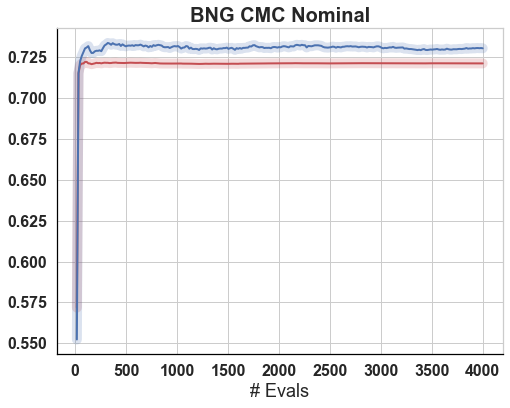}

\includegraphics[width=.34\textwidth]{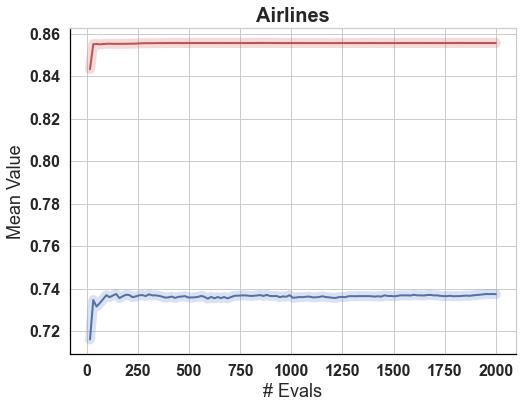}\hfill
\includegraphics[width=.33\textwidth]{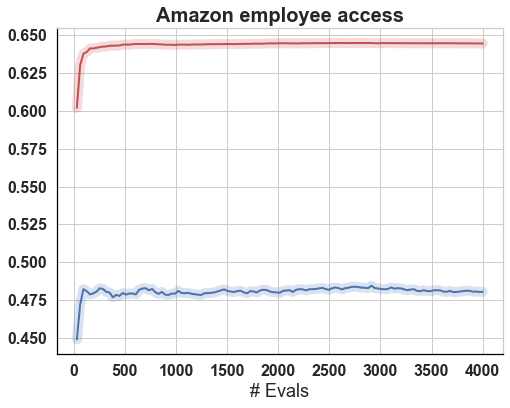}\hfill
\includegraphics[width=.33\textwidth]{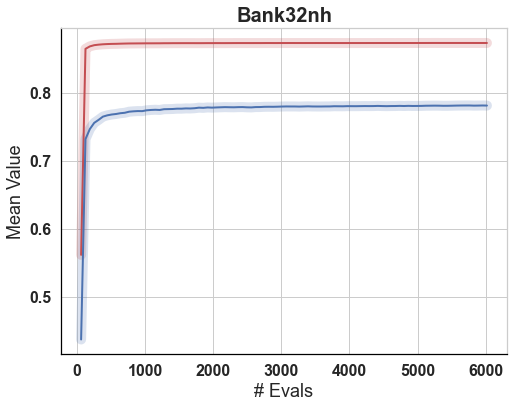}

\includegraphics[width=.34\textwidth]{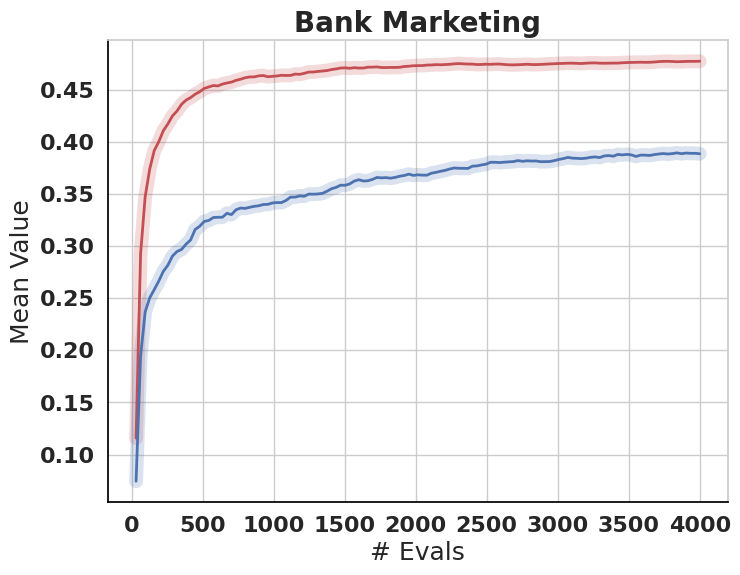}\hfill
\includegraphics[width=.33\textwidth]{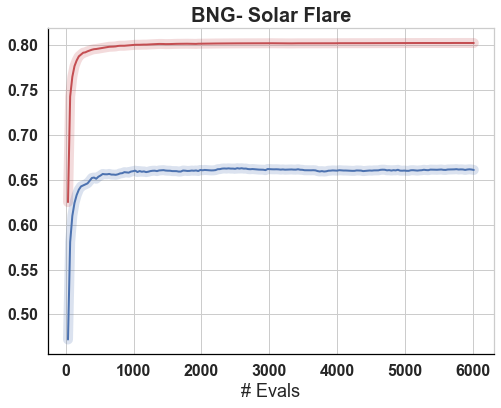}\hfill
\includegraphics[width=.33\textwidth]{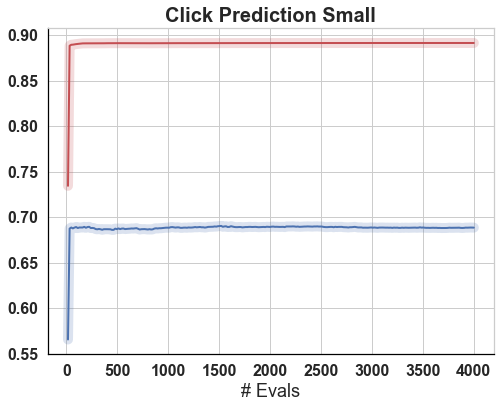}

\caption{\textbf{Comparison between the approximations generated by KernelSHAP and the importance scores obtained from IGNetH.} We assess the similarity of KernelSHAP's approximations to the scores produced by IGNetH during the iterations of data sampling and evaluation by KernelSHAP. It becomes evident that KernelSHAP shows improved accuracy in approximating the scores derived from IGNetH with further data sampling.}
\label{fig:transparency_full1}

\end{figure*}

\begin{figure*}[h]

\centering
\includegraphics[width=.34\textwidth]{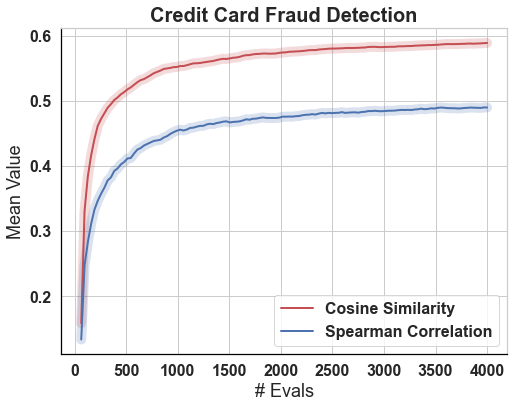}\hfill
\includegraphics[width=.33\textwidth]{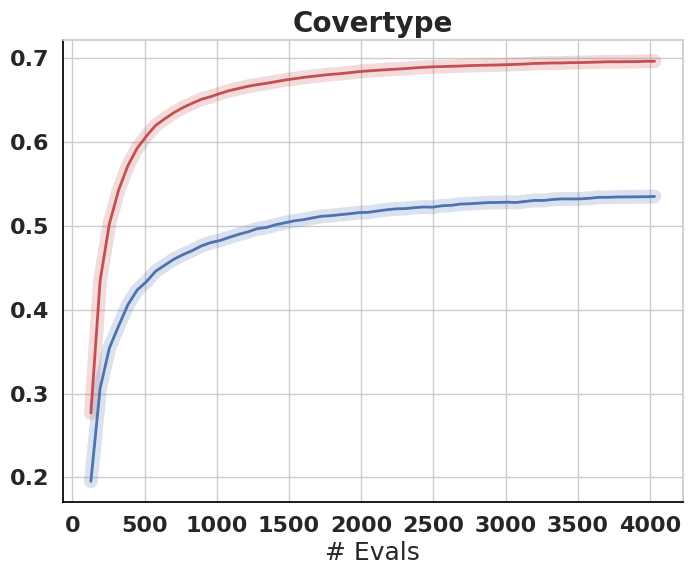}\hfill
\includegraphics[width=.33\textwidth]{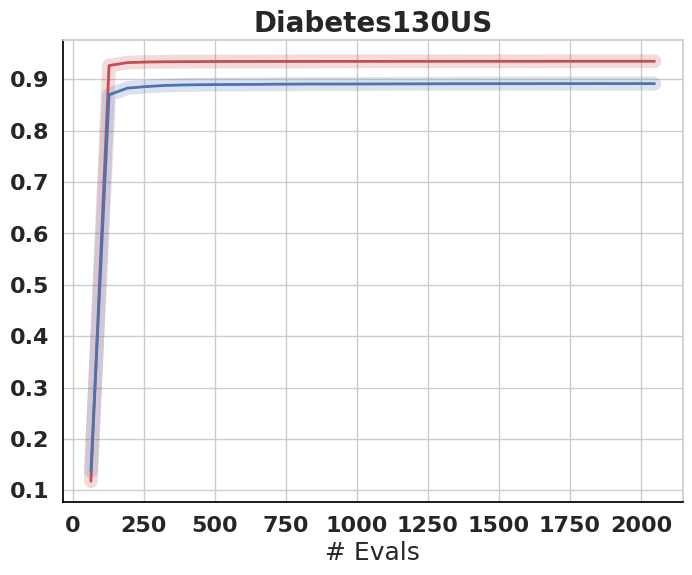}

\includegraphics[width=.34\textwidth]{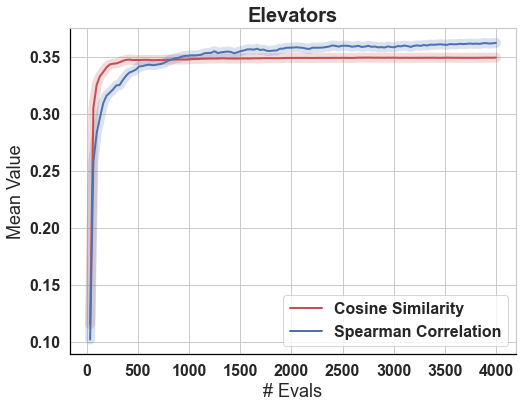}\hfill
\includegraphics[width=.33\textwidth]{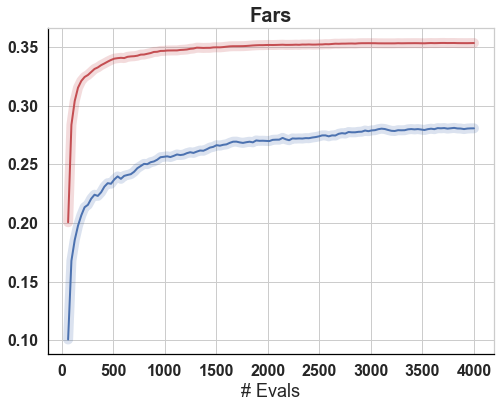}\hfill
\includegraphics[width=.33\textwidth]{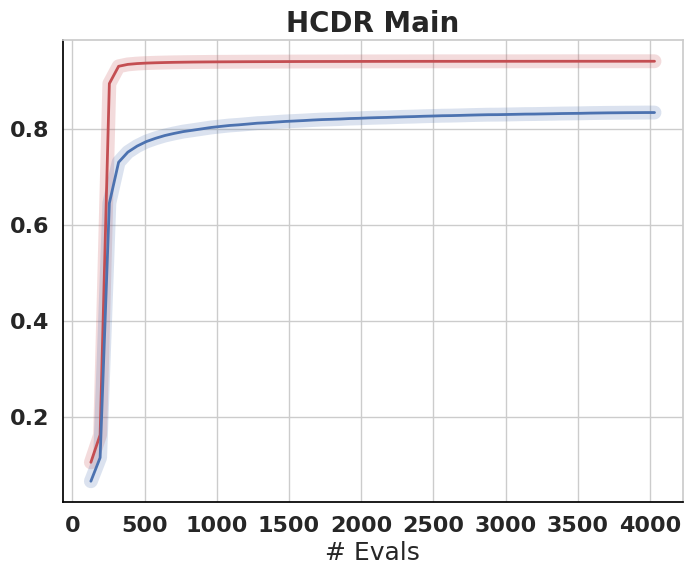}

\includegraphics[width=.34\textwidth]{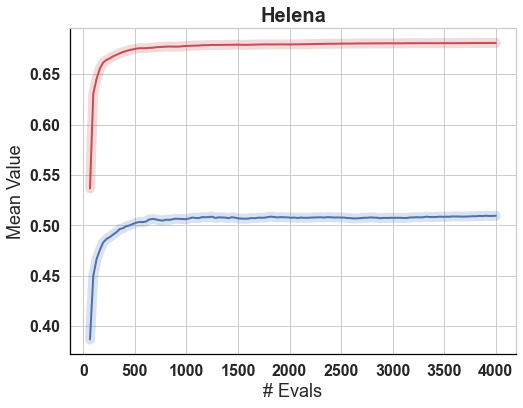}\hfill
\includegraphics[width=.33\textwidth]{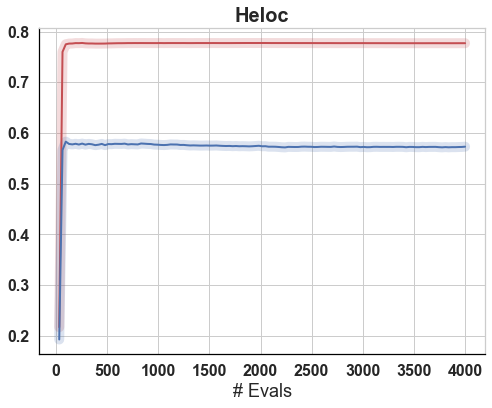}\hfill
\includegraphics[width=.33\textwidth]{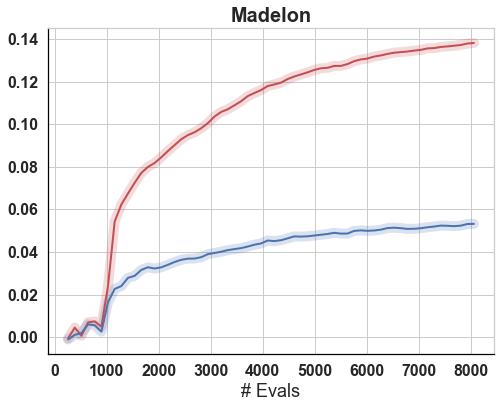}

\includegraphics[width=.34\textwidth]{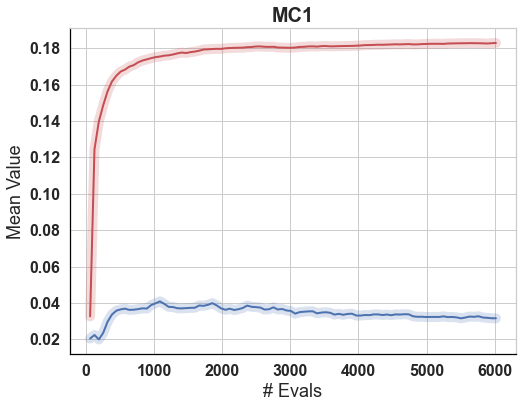}\hfill
\includegraphics[width=.33\textwidth]{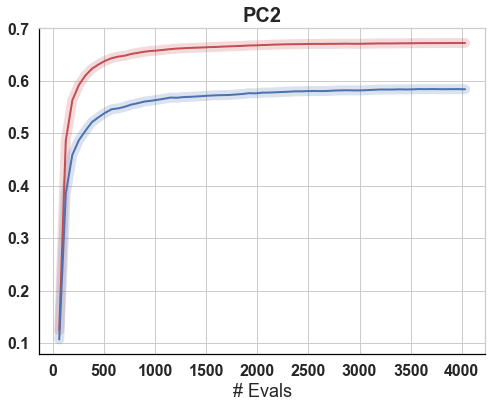}\hfill
\includegraphics[width=.33\textwidth]{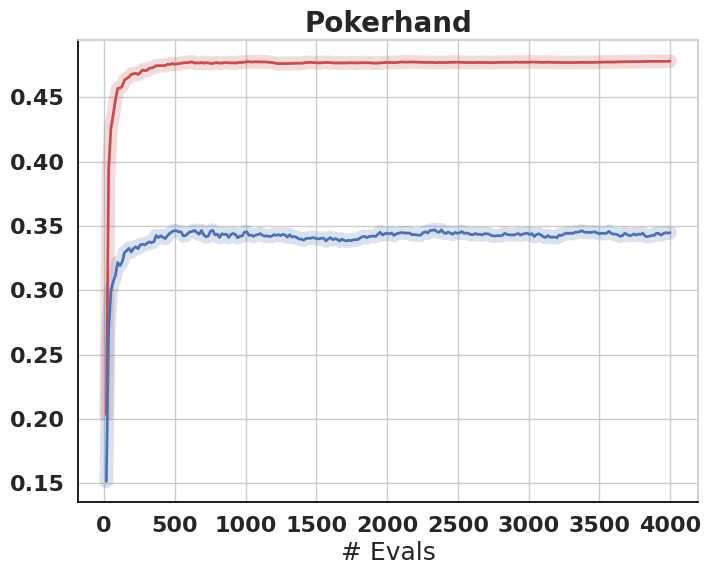}

\includegraphics[width=.34\textwidth]{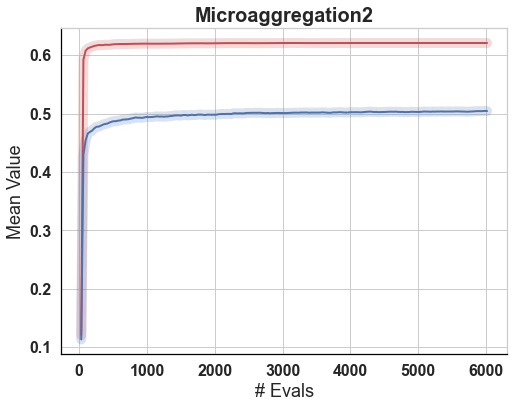}\hfill
\includegraphics[width=.33\textwidth]{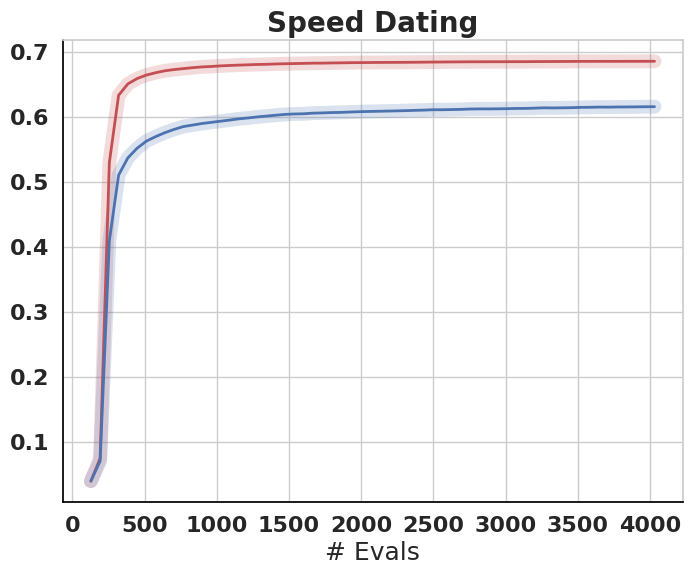}\hfill
\includegraphics[width=.33\textwidth]{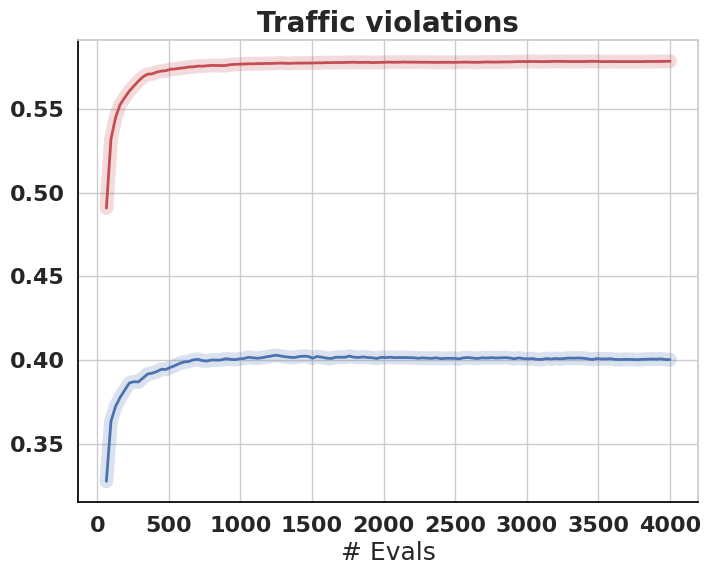}

\caption{\textbf{Comparison between the approximations generated by KernelSHAP and the importance scores obtained from IGNetH.} We assess the similarity of KernelSHAP's approximations to the scores produced by IGNetH during the iterations of data sampling and evaluation by KernelSHAP. It becomes evident that KernelSHAP shows improved accuracy in approximating the scores derived from IGNetH with further data sampling.}
\label{fig:transparency_full2}

\end{figure*}

\section{Specifications of the Hardware and Details of the Datasets}

Table 2 provides details about the datasets used in the experiments, including the number of classes (\# Classes), the number of features (\# Features), the size of the dataset, the size of the training, validation, and test splits, in addition to the weight of the self-loop (SL Wt.). We also provide the number of training iterations and the ID of each dataset on OpenML. 

All the experiments have been conducted in a Python environment on a machine with Intel(R) Core(TM) i9-10885H CPU @ 2.40GHz system and 64.0 GB of RAM. The employed GPUs are NVIDIA® GeForce® GTX 1650 Ti with 4 GB GDDR6 and NVIDIA® GeForce® GTX 1080 Ti with 8 GB. The Python package dependencies are listed with the source code.

\begin{table*}
\caption{The dataset information.}
\centering
\begin{adjustbox}{width=1\textwidth}
\small
\begin{threeparttable}
\begin{tabular}{l c c c c c c c c c}
    \toprule
    \rowcolor[HTML]{EFEFEF} 

    \multicolumn{1}{c}{{\cellcolor[HTML]{EFEFEF}Dataset}} & \multicolumn{1}{l}{\cellcolor[HTML]{EFEFEF}\# Classes} & \# Features & Dataset Size & Training Set & val. Set & Test Set & SL Wt. & Epochs & OpenML ID    \\
    \cmidrule(lr){1-10}
    Abalone & 2 & 8 & 4177 & 2506 & 836 & 835 & 45 & 173 & 720 \\ 
    Airlines & 2 & 7 & 539,383 & 512,413 & 13,485 & 13,485 & 6 & 134 & 1169\\ 
    Amazon employee access & 2 & 9 & 32,769 & 24,576 & 4,097 & 4,096 & 2 & 4 & 4135 \\ 
    Autos & 7 & 25 & 1,000,000 & 950,000 & 25,000 & 25,000 & 12 & 188 & 45080 \\ 
    Bank32nh & 2 & 32 & 8,192 & 5,734 & 1,229 & 1,229 & 4 & 189 & 833 \\ 
    Bank Marketing & 2 & 16 & 45,211 & 33,908 & 5,652 & 5,651 & 6 & 57 & 1461 \\ 
    BNG solar flare & 2 & 12 & 663,552 & 630,374 & 16,589 & 16,589 & 10 & 27 & 1178 \\ 
    BNG cmc nominal & 3 & 9 & 55,296 & 44,236 & 5,530 & 5,530 & 3 & 178 & 119 \\ 
    Click prediction small & 2 & 9 & 39,948 & 27,963 & 5,993 & 5,992 & 10 & 185 & 1220 \\ 
    Covertype & 7 & 54 & 581,012 & 524,362 & 27,599 & 29,051 & 10 & 600 & 1596 \\ 
    Credit Card Fraud Detection & 2 & 30 & 284,807 & 270,566 & 7,121 & 7,120 & 6 & 30 & 42175 \\ 
    Diabetes130US & 3 & 47 & 101,766 & 81,412 & 10,177 & 10,177 & 20 & 20 & 45069 \\ 
    Dota2 Games Results & 2 & 116 & 102,944 & 82,355 & 10,295 & 10,294 & 40 & 78 & 45563 \\ 
    Elevators & 2 & 18 & 16,599 & 11,619 & 2,490 & 2,490 & 8 & 138 & 846 \\ 
    Fars & 8 & 29 & 100,968 & 80,774 & 10,097 & 10,097 & 15 & 286 & 40672 \\ 
    HCDR Main & 2 & 120 & 307,511 & 276,759 & 15,376 & 15,376 & 200 & 20 & 45567 \\ 
    Helena & 100 & 27 & 65,196 & 41,724 & 10,432 & 13,040 & 10 & 489 & 41169 \\ 
    Heloc & 2 & 22 & 10,000 & 7,500 & 1,250 & 1,250 & 20 & 163 & 45023 \\ 
    JM1 & 2 & 21 & 10,885 & 8,708 & 1,089 & 1,088 & 140 & 40 & 1053 \\ 
    Kick & 2 & 32 & 72,983 & 54,737 & 9,123 & 9,123 & 10 & 8 & 41162 \\ 
    Madelon & 2 & 500 & 2,600 & 1,560 & 520 & 520 & 3 & 73 & 1485 \\ 
    MC1 & 2 & 38 & 9,466 & 7,478 & 994 & 994 & 80 & 65 & 1056 \\ 
    Microaggregation2 & 5 & 20 & 20,000 & 12,800 & 3,200 & 4,000 & 20 & 590 & 41671 \\ 
    Numerai28.6 & 2 & 21 & 96,320 & 86,688 & 4,816 & 4,816 & 30 & 29 & 23517 \\ 
    Pokerhand & 10 & 10 & 1,025,009 & 768,756 & 128,127 & 128,126 & 0.6 & 100 & 1567 \\ 
    SF Police Incidents & 2 & 6 & 538,638 & 484,774 & 26,932 & 26,932 & 4 & 16 & 42344 \\ 
    Traffic Violations & 3 & 20 & 70,340 & 52,755 & 8,793 & 8,792 & 8 & 20 & 42345 \\ 
    PC2 & 2 & 36 & 5,589 & 3,353 & 1,118 & 1,118 & 60 & 18 & 1069 \\ 
    Porto Seguro & 2 & 37 & 595,212 & 565,451 & 14,881 & 14,880 & 20 & 12 & 42206 \\ 
    Speed Dating & 2 & 120 & 8,378 & 5,864 & 1,257 & 1,257 & 45 & 20 & 40536 \\ \bottomrule
\end{tabular}
\end{threeparttable}
\end{adjustbox}
\label{table:datasets}
\end{table*}
\end{document}